\documentclass{article}
\usepackage{ijcai26}

\usepackage{times}
\usepackage{soul}
\usepackage[hidelinks]{hyperref}
\usepackage[utf8]{inputenc}
\usepackage[small]{caption}
\usepackage{graphicx}
\usepackage{amsthm}
\usepackage[switch]{lineno}
\newtheorem{example}{Example}

\usepackage[T1]{fontenc}
\usepackage{url}
\usepackage{tikz}
\usepackage{booktabs} 
\usepackage{xcolor}   
\usepackage{orcidlink}
\definecolor{fgred}{HTML}{CD5C5C} 

\usetikzlibrary{shapes.geometric, positioning}
\usepackage{pgfplots}
\pgfplotsset{compat=1.18}
\usepackage{float}
\usepackage{ifthen}
\usepackage{amsmath}
\usepackage{amssymb}
\usepgfplotslibrary{statistics}
\usepackage{pgfplotstable}
\usetikzlibrary{positioning}
\usepackage[ruled,vlined,linesnumbered]{algorithm2e}
\DontPrintSemicolon
\usepackage{graphicx}
\usepackage{xspace}
\usepackage{subcaption}

\usepackage{todonotes}

\newcommand{\mdd}{\textsc{Mdd}\xspace}
\newcommand{\mdds}{\textsc{Mdd}s\xspace}
\newcommand{\nomdd}{\textsc{No-Mdd}\xspace}

\newcommand{\nooverlap}{No-Overlap\xspace}

\newcommand{\earliest}[1]{E^\downarrow_{#1}}
\newcommand{\latest}[1]{L^\uparrow_{#1}}
\newcommand{\orig}[1]{{#1}.orig}
\newcommand{\dest}[1]{{#1}.dest}
\newcommand{\lbl}[1]{\phi_{#1}}

\title{Towards Bound Consistency for the No-Overlap Constraint Using MDDs}

\author{
Amaury Guichard$^1$\orcidlink{0009-0004-2353-1522}
\and
Laurent Michel$^2$\orcidlink{0000-0001-7230-7130}
\and
Hélène Verhaeghe$^1$\orcidlink{0000-0003-0233-4656}
\And
Pierre Schaus$^1$\orcidlink{0000-0002-3153-8941}\\
\affiliations
$^1$UCLouvain, ICTEAM, Louvain-la-Neuve, Belgium\\
$^2$University of Connecticut, Storrs, CT, USA\\
\emails
amaury.guichard@uclouvain.be,
ldm@engr.uconn.edu,
helene.verhaeghe@uclouvain.be,
pierre.schaus@uclouvain.be
}

\begin{document}

\maketitle

\begin{abstract}
Achieving bound consistency for the \nooverlap constraint is known to be NP-complete.
Therefore, several polynomial-time tightening techniques, such as edge finding, not-first-not-last reasoning, and energetic reasoning, have been introduced for this constraint.
In this work, we derive the first bound-consistent algorithm for the \nooverlap constraint.
By building on the \nooverlap \mdd defined by Ciré and van Hoeve, we extract bounds of the time window of the jobs, allowing us to tighten start and end times in time polynomial in the number of nodes of the \mdd.
Similarly, to bound the size and time-complexity, we limit the width of the \mdd to a threshold, creating a relaxed \mdd that can also be used to relax the bound-consistent filtering.
Through experiments on a sequencing problem with time windows and a just-in-time objective ($1 \mid r_j, d_j, \bar{d}_j \mid \sum E_j + \sum T_j$), we observe that the proposed filtering, even with a threshold on the width, achieves a stronger reduction in the number of nodes visited in the search tree compared to the previously proposed precedence-detection algorithm of Ciré and van Hoeve.
The new filtering also appears to be complementary to classical propagation methods for the \nooverlap constraint, allowing a substantial reduction in both the number of nodes and the solving time on several instances.
\end{abstract}

\section{Introduction}

Let $J = \{1, \dots, n\}$, a set of $n$ jobs. Each job $i \in J$ has a release date $r_i$ (i.e., earliest time the task can start), a processing time $p_i > 0$, and a strict deadline $\bar{d}_i > r_i$ (i.e., latest time the task have to end). The disjunctive scheduling problem is the problem of determining start times so that all jobs execute within their time windows without overlapping\footnote{Notations are borrowed from \cite{graham1979optimization}.}. 
This problem is NP-complete \cite{garey1979}. Disjunctive scheduling problems have already been widely studied in the Constraint Programming (CP) community \cite{blaisdisjunctive,green2024using,grimes2015solving,hebrard2025disjunctive,Laborie2018,siala2015two,simonin2015scheduling}.
In CP, this problem is traditionally modeled using start-time variables $s_i \in [r_i, \bar{d}_i - p_i]$ and end-time variables $e_i = s_i + p_i$, or by using interval variables \cite{laborie2008reasoning}. The non-overlapping requirement is enforced by a set of constraints that together define the semantics of the global \nooverlap (also known as Disjunctive) constraint:
$$\forall i,j \in J,\ \mathrm{s.t.}\ i\not= j,\ e_i \leq s_j\ \lor\ e_j \leq s_i$$

Propagators prune inconsistent values by updating the earliest start $\underline{s_i}$ (i.e., lower bound of $s_i$) and latest completion times $\overline{e_i}$ (i.e., upper bound of $e_i$), tightening the time windows. Achieving bound consistency is at least as hard as solving the problem itself, which is NP-complete.
Therefore, many polynomial-time algorithms (most of them running in $O(n^2)$ or $O(n\log(n))$) update the bounds using relaxed reasoning.
The most common approaches are based on edge-finding, not-first/not-last, or detectable precedence reasoning (a non-exhaustive list of such algorithms includes \cite{baptiste2001constraint,carlier1994adjustment,fahimi2018linear,fahimi2014linear,Vilm2005}).

\begin{figure*}[t]
    \centering
    \definecolor{fgligthgray}{HTML}{BDBDBD}
    \definecolor{fgdarkgray}{HTML}{424242}
    \definecolor{gridgray}{HTML}{E0E0E0}
    \definecolor{fgred}{HTML}{CD5C5C} 

    \begin{minipage}[c]{0.25\textwidth}
        \centering
        \sffamily 
        \setlength{\tabcolsep}{5pt} 

        \begin{tabular}{l|rrr}
            \toprule
            Domains & $r_i$ & $\bar{d}_i$ & $p_i$ \\
            \hline
            $t_1$ & 4 & 6 & 2 \\
            $t_2$ & 0 & 10 & 3 \\
            $t_3$ & 0 & 9 & 2 \\
            $t_4$ & \textcolor{fgred}{\textbf{7}} & 19 & 6 \\
            \bottomrule
        \end{tabular}
    \end{minipage}%
    \hfill
    \begin{minipage}[c]{0.74\textwidth}
    \centering
        \begin{tikzpicture}[x=0.4cm, y=0.33cm, font=\small\sffamily,scale=1.6]

            \newcommand{\drawtaskbars}[5]{
                \fill[fgligthgray] (#2, #1) rectangle (#3, #1+0.8);
                \fill[fgdarkgray]  (#4, #1) rectangle (#5, #1+0.8);
            }

            \foreach \i in {0,...,19} {
                 \draw[gridgray, thin] (\i, -3) -- (\i, 1); 
                 \node[above, anchor=south] at (\i, 1) {\scriptsize \i};
            }

            \foreach \y in {0,-1,-2,-3} {
                \draw[gridgray, thin] (0, \y) -- (19, \y);
            }

            \drawtaskbars{0}{4}{6}{4}{6}

            \drawtaskbars{-1}{0}{10}{3}{6}

            \drawtaskbars{-2}{0}{9}{3}{5}

            \fill[fgred]       (7, -3) rectangle (8, -3+0.8); 
            \fill[fgligthgray] (8, -3) rectangle (19, -3+0.8); 
            \fill[fgdarkgray]  (10, -3) rectangle (16, -3+0.8); 

            \node[white] at (5,0.4) {\scriptsize $t_1$};
            \node[white] at (4.5,-0.6) {\scriptsize $t_2$};
            \node[white] at (4,-1.6) {\scriptsize $t_3$};
            \node[white] at (13,-2.6) {\scriptsize $t_4$};
        \end{tikzpicture}
    \end{minipage}
    \caption{Job domains and timeline visualization with tightened start time for $t_4$.}
    \label{fig:job_domains_red}
\end{figure*}
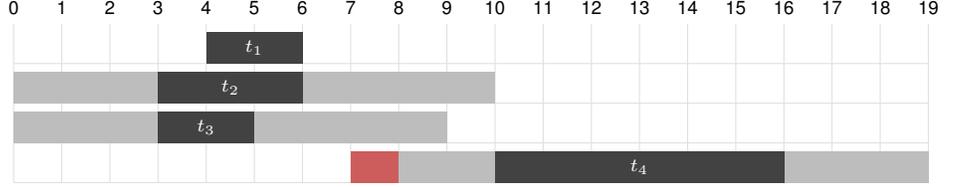

\begin{example}
\label{ex:scheduling}
A small \nooverlap instance is given in Fig.~\ref{fig:job_domains_red}.
Job $t_4$ cannot start at time $7$, as $t_2$ and $t_3$ cannot both be scheduled before $t_1$, nor can they be after $t_4$. 
Ordering $\langle t_2, t_1, t_3, t_4 \rangle$ allows for the earliest feasible start time of $8$ for $t_4$.
None of the common filtering techniques can detect that $t_4$ cannot start at time $7$. In contrast, a bound-consistent filtering algorithm would detect it.
\end{example}

To the best of our knowledge, a guaranteed bound-consistent (BC) filtering has not yet been developed. 
Despite a worst-case exponential time complexity, it could be valuable. First, it would allow researchers to check filtering properties, such as in \cite{gillard2019solvercheck} 
and measure the filtering gap of the existing polynomial-time algorithm to guide future research on filtering algorithms for the \nooverlap.
It would also allow testing whether, for a given problem, increasing its filtering to BC significantly reduces the number of nodes (e.g., using the replay of \cite{VanCauwelaert2015}).

Ciré et al. \cite{Cire2012} introduced a Multi-valued Decision Diagram (\mdd) based on job sequencing, denoted \nomdd hereafter. Each edge is labeled with a job. Consequently, a node at layer $k$ corresponds 
to a state where a prefix of $k$ jobs has been sequenced. Since the width of the \nomdd can grow rapidly (proportional to the number of permutations), they propose to bound it to $W$ via node-merging (i.e., creating a relaxed \nomdd). They also infer job precedence relations from it to filter start times. Given a built \nomdd, detecting all precedences takes $O(n^3 W)$ time,
or $O(n^2 |\mathcal{V}|)$ (with $\mathcal{V}$ the set of nodes).

\emph{The purpose of this paper, building on 
the \nomdd \cite{Cire2012}, is threefold. It offers stronger propagation when working with exact \mdds. It proposes a practical polynomial filtering based on relaxed \mdds. And it delivers an empirical evaluation.}  
Specifically, it first describes a BC filtering
that inspects all edges of the exact \nomdd{} in ${\cal O}(|\mathcal{E}|)$ time (with $\mathcal{E}$, the set of edges) for a  filtering stronger than~\cite{Cire2012}.
It then presents a \emph{polynomial-time} propagator using a relaxed \nomdd
running in ${\cal O}(n^2W)$, whose implementation takes advantage of the dynamic refinements found in Haddock~\cite{Gentzel2020}. 
Finally, it presents an empirical evaluation where the new propagator is a redundant constraint. The evaluation assesses the filtering strength relative to the state of the art and the precedence-based method \cite{Cire2012}.

\section{Bound-Consistent \nooverlap Filtering}

We now recall the \nooverlap \mdd from~\cite{Cire2012} and proceed to the first contribution: achieving BC when filtering start times with an exact \mdd. 

\subsection{The \nooverlap Exact \mdd}

\begin{figure*}[t!]
    \centering
    \begin{subfigure}{0.68\textwidth}
        \centering
        \begin{tikzpicture}[scale=1,
    node/.style={
        draw,
        thick,
        rounded corners=5pt,
        minimum height=4mm,
        minimum width=15mm,
        inner sep=3pt,
        align=center
    },
    edge label/.style={left, fill=white, inner sep=2pt}
]

\def\layerheight{1.3}  
\def\layerwidth{2.2}     

\node[node] (x1) at (0, 0)
{\small $\langle A^\downarrow, \earliest{}\rangle=\langle\emptyset,0\rangle$};

\node[node, draw=red] (x9) at (-0.5*\layerwidth, -1*\layerheight)
{\small $\langle\{4\},13\rangle$};
\node[node] (x2) at (0.5*\layerwidth, -1*\layerheight)
{\small $\langle\{2\},3\rangle$};
\node[node] (x3) at (1.5*\layerwidth, -1*\layerheight)
{\small $\langle\{3\},2\rangle$};
\node[node, draw=orange] (x10) at (-1.5*\layerwidth, -1*\layerheight)
{\small $\langle\{1\},6\rangle$};

\node[node] (x4) at (0.5*\layerwidth, -2*\layerheight)
{\small $\langle\{1,2\},6\rangle$};
\node[node] (x5) at (1.5*\layerwidth, -2*\layerheight)
{\small $\langle\{1,3\},6\rangle$};
\node[node, draw=red] (x11) at (-0.5*\layerwidth, -2*\layerheight)
{\small $\langle\{1,2\},9\rangle$};
\node[node, draw=red] (x12) at (-1.5*\layerwidth, -2*\layerheight)
{\small $\langle\{1,3\},8\rangle$};
\node[node, draw=red] (x13) at (-2.5*\layerwidth, -2*\layerheight)
{\small $\langle\{1,4\},13\rangle$};

\node[node] (x6) at (0.5*\layerwidth-0.05, -3*\layerheight)
{\small $\langle\{1,2,3\},8\rangle$};
\node[node] (x7) at (1.5*\layerwidth, -3*\layerheight)
{\small $\langle\{1,2,3\},9\rangle$};

\node[node] (x8) at (0, -4*\layerheight)
{\small $\langle J,H\rangle$};

\draw[->] (x1) -- node[edge label] {\small 1} (x10);
\draw[->] (x1) -- node[edge label] {\small 2} (x2);
\draw[->] (x1) -- node[edge label] {\small 3} (x3);
\draw[->] (x1) -- node[edge label] {\small 4} (x9);

\draw[->] (x2) -- node[edge label] {\small 1} (x4);
\draw[->] (x3) -- node[edge label] {\small 1} (x5);
\draw[->] (x4) -- node[edge label] {\small 3} (x6);
\draw[->] (x5) -- node[edge label] {\small 2} (x7);
\draw[->] (x6) -- node[edge label] {\small 4} (x8);
\draw[->] (x7) -- node[edge label] {\small 4} (x8);

\draw[->] (x10) -- node[edge label] {\small 2} (x11);
\draw[->] (x10) -- node[edge label] {\small 3} (x12);
\draw[->] (x10) -- node[edge label] {\small 4} (x13);

\end{tikzpicture}
        \caption{Exact \nomdd with dead ends nodes yet to be pruned}
        \label{fig:mdd_exact_complete}
    \end{subfigure}\hfill
    \begin{subfigure}{0.3\textwidth}
        \centering
        \begin{tikzpicture}[scale=1,
    node/.style={
        draw,
        thick,
        rounded corners=5pt,
        minimum height=4mm,
        minimum width=15mm,
        inner sep=3pt,
        align=center
    },
    edge label/.style={left, fill=white, inner sep=2pt}
]

\def\layerheight{1.3}  
\def\layerwidth{2.5}     

\node[node] (x1) at (0.5*\layerwidth, 0)
{\small $\langle A^\downarrow, \earliest{}\rangle=\langle\emptyset,0\rangle$};

\node[node] (x2) at (0, -1*\layerheight)
{\small $\langle\{2\},3\rangle$};
\node[node] (x3) at (\layerwidth, -1*\layerheight)
{\small $\langle\{3\},2\rangle$};

\node[node] (x4) at (0, -2*\layerheight)
{\small $\langle\{1,2\},6\rangle$};
\node[node] (x5) at (\layerwidth, -2*\layerheight)
{\small $\langle\{1,3\},6\rangle$};

\node[node] (x6) at (0, -3*\layerheight)
{\small $\langle\{1,2,3\},8\rangle$};
\node[node] (x7) at (\layerwidth, -3*\layerheight)
{\small $\langle\{1,2,3\},9\rangle$};

\node[node] (x8) at (0.5*\layerwidth, -4*\layerheight)
{\small $\langle J,H\rangle$};

\draw[->] (x1) -- node[edge label] {\small 2} (x2);
\draw[->] (x1) -- node[edge label, right] {\small 3} (x3);
\draw[->] (x2) -- node[edge label] {\small 1} (x4);
\draw[->] (x3) -- node[edge label, right] {\small 1} (x5);
\draw[->] (x4) -- node[edge label] {\small 3} (x6);
\draw[->] (x5) -- node[edge label, right] {\small 2} (x7);
\draw[->] (x6) -- node[edge label] {\small 4} (x8);
\draw[->] (x7) -- node[edge label, right] {\small 4} (x8);

\end{tikzpicture}
        \caption{Exact \nomdd}
        \label{fig:mdd-exact}
    \end{subfigure}
    \caption{Construction process of an exact \nomdd}
    \label{fig:mdd-exact-generation}
\end{figure*}
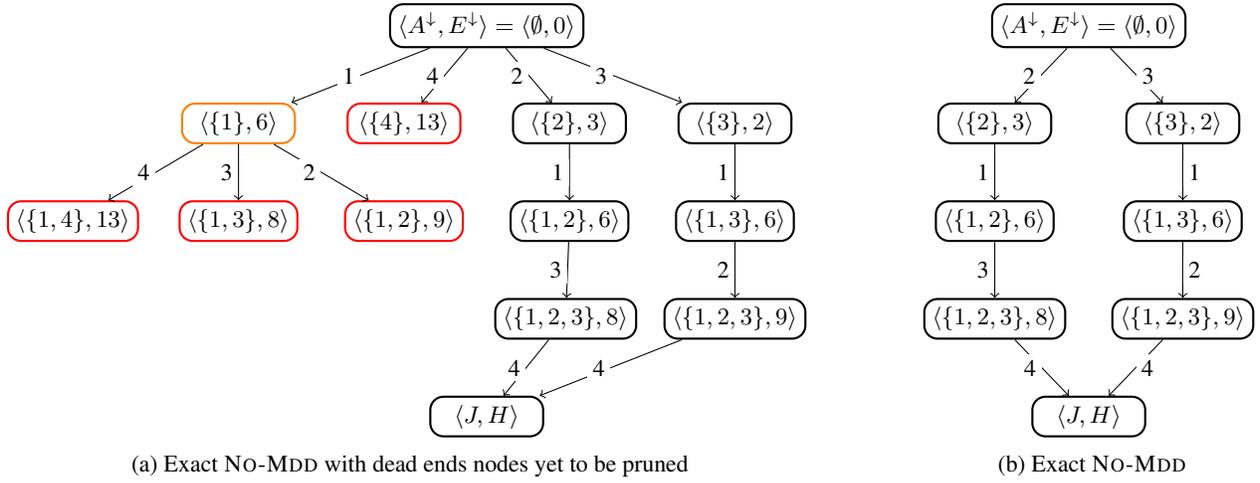

An \mdd can be specified by a set of \textit{states} $\mathcal{S}$, a set of possible \textit{labels} $\mathcal{U}$, a \textit{label} generating function $\lambda : \mathcal{S} \to \mathcal{U}$ (i.e., possible edges from a state), and a \textit{state transition} function $\tau : \mathcal{S} \times \mathcal{U} \to \mathcal{S}$ computing the new state for a state and a label. For the \nomdd, these are:
\begin{itemize}
    \item A state is a tuple $\langle A^{\downarrow}, \earliest{}\rangle$, where $A^\downarrow \subseteq  J$ is the set of jobs already placed, and $\earliest{}$ is the earliest time at which the next job can be placed. The initial state (\textit{root}) is $\langle \emptyset, 0\rangle$, i.e., no activities sequenced yet. The target state (\textit{sink}) is defined as $\langle J, H \rangle$, i.e.,  the unique state where all the jobs have been sequenced, and $H$ is an upper-bound on the latest completion time (horizon).
    \item Labels consist of the jobs to be scheduled (i.e., $\mathcal{U} = J$).
    \item $\lambda(\langle A^{\downarrow}, \earliest{} \rangle) = J \setminus ( A^\downarrow \cup \{  i \in J \mid \max(\earliest{}, \underline{s}_i) + p_i > \overline{e}_i \} )$.
    This function filters out the jobs that have already been sequenced earlier and the ones that cannot be scheduled while satisfying their deadlines.  
    \item The state transition fuction $\tau$ is defined as: If $A^{\downarrow} \cup i = J$, $\tau(\langle A^{\downarrow}, \earliest{}\rangle , i) = \langle J, H\rangle$, otherwise, $\tau(\langle A^{\downarrow}, \earliest{}\rangle , i) = \langle A^{\downarrow} \cup i, \max(\earliest{}, \underline{s}_i) + p_i\rangle$.
This function ensures that job $i$ is never scheduled earlier than the earliest feasible time of the originating state, $\earliest{}$, nor than its own.
\end{itemize}
The \nomdd is compiled top-down layer-wise, ensuring that each state is unique in each layer. One can note that applying $\lambda$ to the \textit{sink} node yields an empty set, ensuring, by construction, an \mdd of at most $|J|$ layers of edges, leading to sequences of at most $|J|$ jobs. Since solutions are, by design, permutations of the $J$ jobs, one knows that nodes on a path that does not reach the \textit{sink} are not part of solutions. This is why a bottom-up pass is then applied to remove every node that does not belong to at least one path to the target.
Let $\mathcal{E}$ denote the set of edges of the \nomdd.
Every directed $e=(u,v) \in \mathcal{E}$ is labeled by job $\lbl{e}$.
We further define $\orig{e}=u$ and $\dest{e}=v$.

\begin{example}
The construction process of the exact \nomdd is shown in Fig.~\ref{fig:mdd-exact-generation}. Figure~\ref{fig:mdd_exact_complete} represents the top-down computations of each layer of the exact \nomdd. Nodes display state $\langle A^\downarrow, \earliest{}\rangle$. In red, one can see the childless nodes that will be removed at the bottom-up pass, and in orange, the ones removed consequently.
The final exact \nomdd is given at Fig.~\ref{fig:mdd-exact}. 
\end{example}

\subsection{Bound-Consistent Filtering using the \nomdd}

The earliest start time of job $i \in J$ is computed as the minimum of
earliest start states among all edges labeled with $i$ in the \nomdd.
More formally:
\begin{equation}
\underline{s}_i \leftarrow 
\max\left(
    \underline{s}_i,\ 
    \min_{\{ e \in \mathcal{E} \mid \lbl{e}=i \}}
        \bigl( \earliest{\orig{e}} \bigr)
\right)\label{eq:tightenstart}
\end{equation}

This is valid because every path from the root to the target node of the \nomdd encodes a feasible sequencing of all jobs in $J$ and guarantees that, by scheduling them sequentially in that order at their earliest start times, the deadline constraints are also satisfied.
Therefore, the formula implicitly takes the earliest start time of job $i$ among all valid sequences.
Therefore, these filtering rules ensure bound-consistency of the \nooverlap constraint.

Note that the filtering can be performed for all jobs in $O(|\mathcal{E}|)$ time, since one can create an array initialized as $\underline{s}' = [+\infty \mid i \in J]$ and update the corresponding entry ${\underline{s}'}_{\lbl{e}} \leftarrow \min\bigl({\underline{s}'}_{\lbl{e}} , \earliest{\orig{e}}\bigr)$ in constant time for each edge $e \in \mathcal{E}$.
A similar filtering can be applied to tighten the latest end times.
Given interval domains for the start- and end-times variables, the fix-point is obtained in one pass.

\begin{example}
    Considering the exact \nomdd in Fig.~\ref{fig:mdd-exact} and job $t_4$,
    two edges have this job as a label. Therefore, the earliest start for this job is $\underline{s}_4 \leftarrow \min(8,9)$.
\end{example}

\section{Relaxed Bound-Consistent \nooverlap}

The complexity for the BC filtering of the \nooverlap constraint is thus linear in the number of edges of the \nomdd. 
However, this number can grow exponentially. 
To palliate that, \cite{Cire2012} suggested working on a relaxed \nomdd, by bounding the width of each layer to $W$, encoding a super-set of all the valid sequences in the exact \nomdd.
This is achieved by merging certain states within layers. This section describes how to build a relaxed \nomdd \cite{Cire2012}. 
This section also explains how to refine the \nomdd, following the Haddock framework \cite{Gentzel2020}.

\subsection{The Relaxed \nooverlap \mdd}

Given an \mdd representing a set of solutions $S$, its relaxation represents a super set of $S$. 
A relaxed \mdd can be obtained by merging nodes with different states on a given layer of an \mdd. 
This reduces its size while adding new invalid paths. 
To construct a top-down relaxed \mdd, one needs to extend the definition of a state to enable node merging, as done in \cite{Cire2012}.
The specification of a relaxed \nomdd including the node-merging operator $\oplus$ is given next.
\begin{itemize}
    \item A state is a tuple  $\langle A^{\downarrow}, S^{\downarrow}_u, \earliest{}, K \rangle$, where $A^\downarrow \subseteq  J$ represents the set of jobs in every path leading to this node, $S^{\downarrow}_u \subseteq A^\downarrow$ is the set of jobs in at least one path leading to this node, and
    $\earliest{}$ is a lower-bound on the earliest start for the next job to be scheduled after all the already placed ones, and $K$ is the number of scheduled jobs (ID of the node layer). The initial state is $\langle \emptyset,\emptyset, 0, 0 \rangle$, a state where no activities have been sequenced yet, and $0$ is assumed to be a lower-bound on the earliest start of all the jobs. The target state is defined as $\langle J, J, H, n \rangle$, that is, the state where all the jobs have been sequenced ($H$ is the horizon, i.e., the upper-bound on the latest completion time). 
    \item $\mathcal{U}=J$.
    \item $ \lambda'(\langle A^{\downarrow}, S^{\downarrow}, \earliest{}, K \rangle) = \lambda(\langle A^{\downarrow}, 
    \earliest{}\rangle) \setminus \Bigl( \mathbf{1}_{\{|S^{\downarrow}| = K\}} S^{\downarrow} \Bigr)
$ where $\mathbf{1}$ is the indicator function to express the conditional set subtraction. This function filters out jobs that have already been sequenced earlier and those that cannot be scheduled while satisfying their deadlines.
    \item The state transition function $\tau'$ is defined as: If $K=n-1$, $\tau'(\langle A^{\downarrow}, S^{\downarrow}, \earliest{},K \rangle , i) = \langle J,J,  H, n \rangle$, otherwise $\tau'(\langle A^{\downarrow}, S^{\downarrow}, \earliest{},K \rangle , i) =\langle A^{\downarrow} \cup \{i\},S^{\downarrow} \cup \{i\}, \max(\earliest{},\, \underline{s}_i) + p_i, K+1 \rangle$.
It ensures that job $i$ is never scheduled earlier than the earliest feasible time of the originating state, $\earliest{}$, nor earlier than its own earliest start time.
\item $\oplus(\langle A_u^{\downarrow}, S_u^{\downarrow}, \earliest{u}, K \rangle, \langle A_v^{\downarrow}, S_v^{\downarrow}, \earliest{v} ,K \rangle) = \langle \langle A_u^{\downarrow} \cap A_v^{\downarrow}, S_u^{\downarrow} \cup S_v^{\downarrow}, \min (\earliest{u},\earliest{v}), K \rangle$

\end{itemize}

The top-down compilation of a relaxed \mdd is similar to that of an exact one except, after expanding a layer, it needs to reduce its size to $W$ by merging nodes ($\oplus$). 
We merge nodes by grouping them in buckets \cite{verhaegheimproved} based on their $\earliest{}$ property. 
Buckets are obtained by partitioning the ranges of values of $\earliest{}$ into $W$ sub-intervals and merging together nodes belonging to the same sub-interval. 

\begin{figure*}[t]
    \centering
    \includegraphics[width=0.8\linewidth]{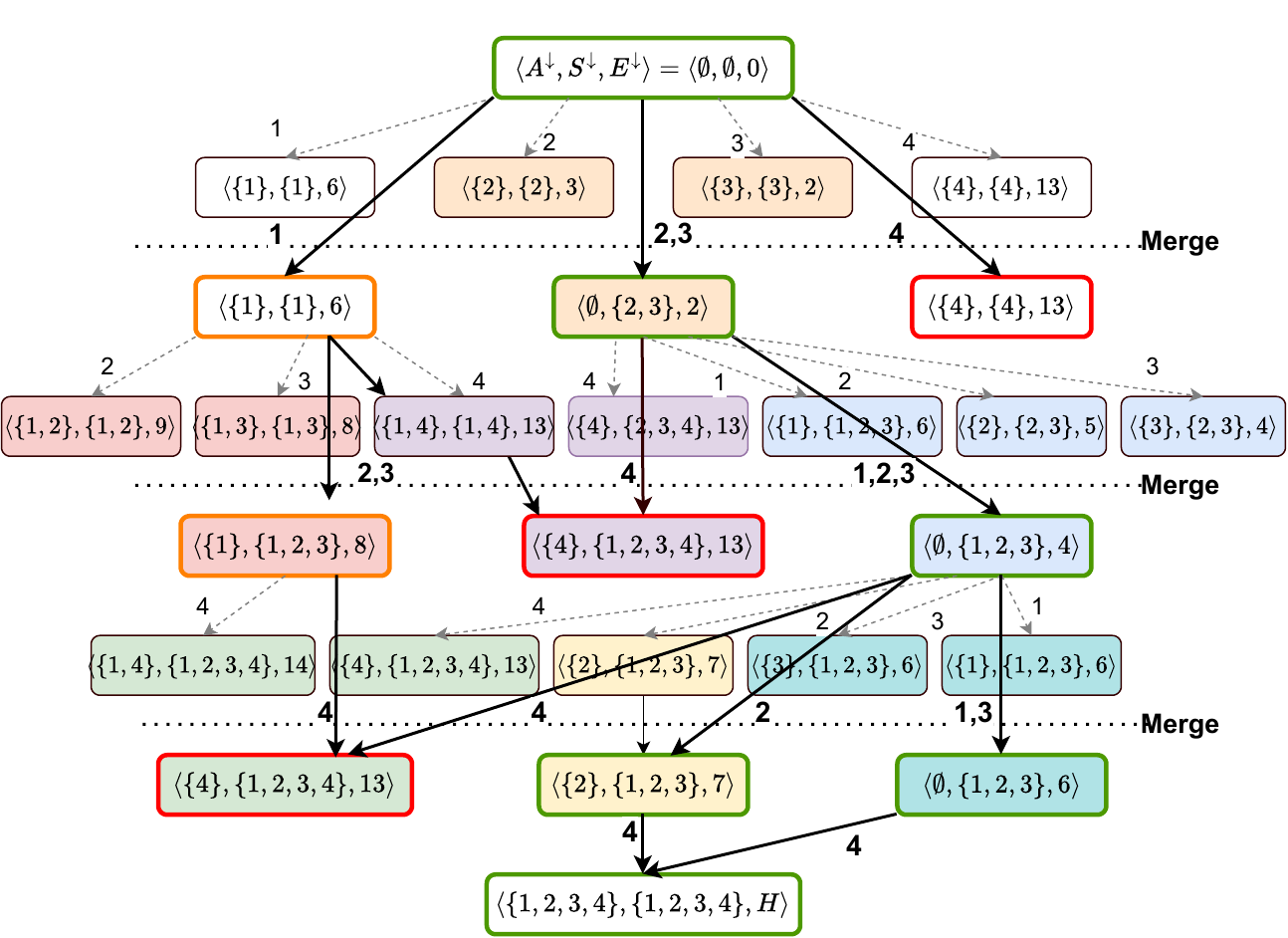}
    \caption{Relaxed \nomdd generation bound to a width of 3}
    \label{fig:3wMDD}
\end{figure*}

\begin{example}
Fig.~\ref{fig:3wMDD} shows the top-down compilation of a relaxed \nomdd. 
Property $K$ (trivial) is omitted for clarity reasons. 
Starting at the root, four nodes are created ($\lambda'$ generates valid outgoing labels, $\tau'$ creates new states). This layer is reduced by bucketing (orange nodes merged).
On the next layer, seven nodes are generated, bucketed, and merged. The process is repeated until each layer is constructed.
Then, the \mdd needs to be made sound again by removing the three dead-end nodes (circled in red) and propagating their removal upward (nodes circled in orange are removed). Circled green nodes composed the final \mdd.
One remark: even though the width is limited to 3 (higher than the exact \mdd{}'s width of 2 in Fig.~\ref{fig:mdd-exact}), the final \mdd is not exact. This is due to the introduction of infeasible solutions during merging, leading to relaxed states that cannot remove some incorrect edges.
Rule~\eqref{eq:tightenstart} does not tighten the earliest start of $t_4$ to 8 yet, but, since the full width is not used, we can refine this \mdd.
\end{example}

\subsection{Incremental update of the relaxed \nomdd}

During the search, along any branch of the search tree, the job time windows can only shrink. This might cause some transitions to become invalid according to $\lambda'$.
One could recompute the \nomdd from scratch, but it turns out to be less costly to update it incrementally using a \emph{refinement} procedure.
It first updates all properties in a top-down manner and deletes transitions when necessary. As a result, some nodes might become orphans or dead ends. After their removal, some layer might be left with fewer nodes than the allowed limit $W$. In such cases, nodes from previous layers are expanded again via the $\tau'$ function, and the layer is compressed when required using the merge operator $\oplus$.
Such refinement procedures are well described in \cite{Cire2012} and \cite{Gentzel2020}, which we adapt to our context here.

\begin{figure*}[t]
    \centering
    \includegraphics[width=0.95\linewidth]{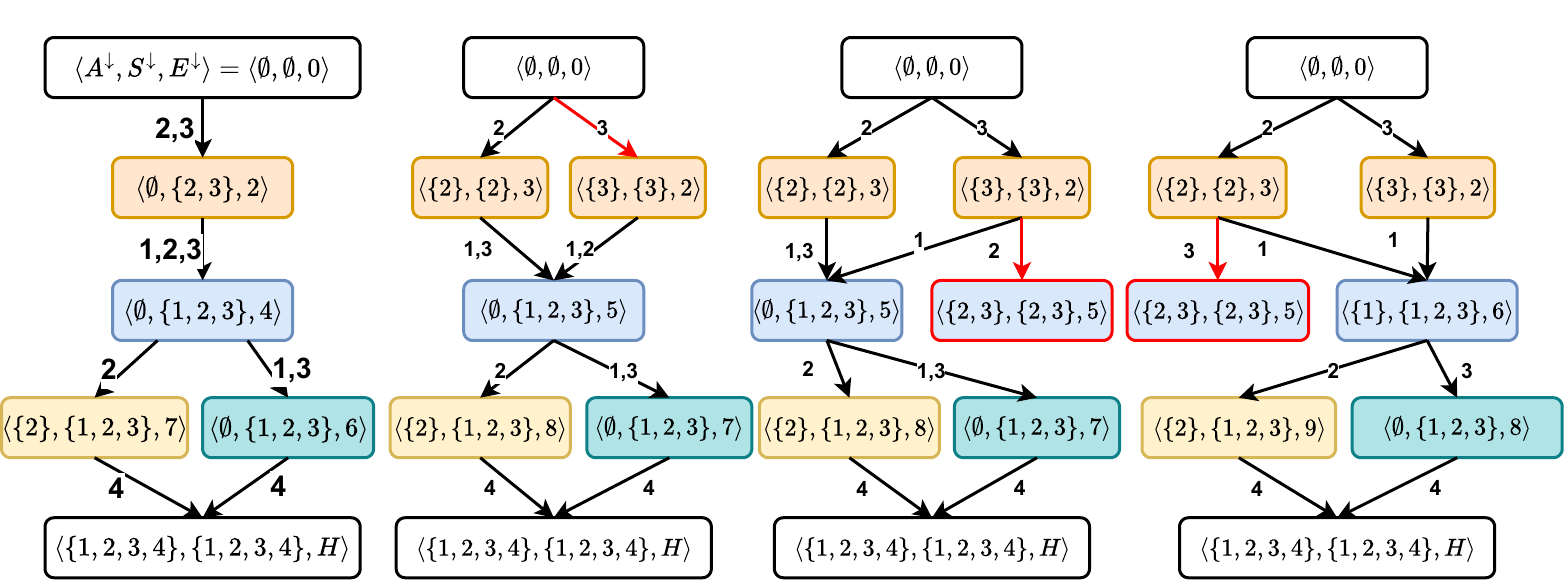}
    \caption{Edges extractions occurring during refinement of the relaxed \mdd}
    \label{fig:refinement}
\end{figure*}

\begin{example}
The refinement process is illustrated in Fig.~\ref{fig:refinement}. From top to bottom, a relaxed node is selected, one of its incoming edges is extracted, and a new node is created to receive it. 
On the example, first, the edge labeled $3$ is extracted from the node $\langle \emptyset,\{2,3\},2\rangle$, and the \mdd is updated. By the filtering rule~\eqref{eq:tightenstart}, the value $7$ is computed.
Then, when extracting the edge labeled $2$ and then the edge labeled $3$ from the node $\langle\emptyset,\{1,2,3\},5\rangle$, the resulting relaxed \mdd is now strong enough to tighten the earliest start time of $t_4$ to~$8$.
\end{example}

So far, we have described only the \nomdd used for filtering the earliest start times of the jobs, and noted that another \mdd could be built in reverse order to prune the latest end times as well.
In practice, as in \cite{Cire2012,Gentzel2020}, we combine these two \mdd by extending the state with the corresponding bottom-up values, i.e., $A^\uparrow$, $S^\uparrow$, $\latest{}$. The update of the states is thus performed both top-down and bottom-up, allowing for a stronger edge check based on the parts of the sequence before and after the edge.

\subsection{BC filtering/precedences extraction on the relaxed \nomdd}

Applying our BC filtering or extracting the precedences on a relaxed \nomdd follows the same formulas as in the exact ones. However, due to the relaxed nature of this \nomdd, the same level of propagation is not guaranteed to be reached. This is why we speak about \textit{Relaxed BC filtering} in that case.


To recall, precedences are extracted from an exact \nomdd by a combination of top-down $A^\downarrow$ and bottom-up $A^\uparrow$ properties of a node (Theorem 6 in \cite{Cire2012}). 
A job $i$ precedes a job $j$ (denoted $i \prec j$) if and only if $\forall\ \mathrm{nodes}\ u, (j \notin A^\downarrow_u) \lor (i \notin A^\uparrow_u)$. Verifying it requires checking each job pair. 
On relaxed \nomdd, the $S^\downarrow$ and $S^\uparrow$ are used to extract the precedence (Corollary 8 in \cite{Cire2012}). A job $i$ precedes a job $j$ (denoted $i \prec j$) if and only if $\forall\ \mathrm{nodes}\ u, (j \notin S^\downarrow_u) \lor (i \notin S^\uparrow_u)$.
The time complexity is thus $O(n^2|\mathcal{V}|)$ (with $\mathcal{V}$ the set of nodes of the \nomdd)\footnote{Can be expressed as $\Omega(n |\mathcal{E}|)$ to ease comparison with the $O( |\mathcal{E}|)$ for the BC filtering} or $O(n^3 W)$, given $W$ the width of the \mdd.


\begin{example}
In the running example, using the exact \mdd, the precedences that can be extracted are $1 \prec 4$, $2 \prec 4$, $3 \prec 4$. Precedence extraction also fails to tighten the time window of $t_4$. Therefore, even on an exact \nomdd, the precedence extraction filtering does not reach the BC one for the \nooverlap.
\end{example}


\section{Experiments}
\label{sec:expe}

We study experimentally the problem of just-in-time scheduling ($1 \mid r_j,d_j, \bar{d}_j \mid \sum E_j + \sum T_j$) \cite{graham1979optimization,baptiste2008lagrangian}. This problem involves scheduling $n$ jobs on a single machine, given their release times $r_j$, processing times $p_j$, deadlines $d_j$, and a strict deadline $\bar{d}_j$ (i.e., the latest completion time).
Given $E_j$ the earliness of job $j$ (i.e., $E_j=\max(0, d_j-e_j)$) and the tardiness $T_j$ (i.e., $T_j=\max(0,e_j-d_j)$), the goal of the problem is to minimize the total sum of the earliness and tardiness of the jobs (i.e., $\sum E_j + \sum T_j$) while scheduling, in a non-overlapping manner, the jobs within their time window defined by their release time and strict deadline. 
We generated a given number of instance of various sizes (depending on the experiment). The processing time of each job was randomly selected from 1 to 25. To generate an instance, jobs were placed sequentially, and each job's time window was defined to overlap with the two preceding and two following jobs. The desired deadline was then randomly selected within this time window.

Our source code is available online\footnote{Anonymized code in the Appendix} and is implemented as a constraint in MaxiCP~\cite{MaxiCP2024}, an extended version of MiniCP~\cite{MiniCP} solver. The experiments were run on an Intel Xeon Platinum 8160 @ 2.100GHz with 96 cores and 320 GB of RAM. 

We compare four models: (i) model (\textit{baseline}) using only all the filtering algorithms from \cite{Vilm2005}\footnote{That is overload-check, detectable precedences, not-first not-last, and edge finding.} for the \nooverlap, (ii) model (\textit{Relaxed BC Filtering}) with the \nomdd, relaxed BC filtering and \nooverlap (redundant), (iii) model (\textit{Precedence Extraction/PE}) with the \nomdd, precedence extraction and \nooverlap, and (iv) model (\textit{BC Filtering/BC}) with the \nomdd, exact BC filtering and \nooverlap (redundant). The last model is not used in all experiments (especially with bigger instances) due to its complexity.

To avoid any bias made by discrepancy in programming language, \mdd refinement strategies, solver used, we implemented the precedence extraction within our solver. This allows both \mdd-based filtering to compare on a similar environment.
Moreover, we use a replay-based search framework \cite{VanCauwelaert2015} to ensure a fair comparison between the models. 
First, the weakest \textit{baseline} model is used with the \emph{conflict ordering search} \cite{Gay2015}) with a time limit of one minute. During this run, the sequence of branching decisions is recorded, thereby defining a reference search tree (possibly partial if the run times out).
This recorded search tree is then replayed for the other models, which differ only in their pruning strength. As a result, all models are forced to explore exactly the same parts of the search tree, and any differences in performance can be attributed solely to the additional filtering they provide by pruning this forced search tree exploration. 
In particular, this avoids confounding effects due to heuristic interactions\footnote{It is well known that stronger propagation combined with first-fail–based heuristics may paradoxically lead to the exploration of larger search trees.}.

\subsection{Search Space Reduction and Runtime Comparison}

\begin{figure}[t!]
\begin{subfigure}{0.49\textwidth}
\centering
    \includegraphics[width=\textwidth]{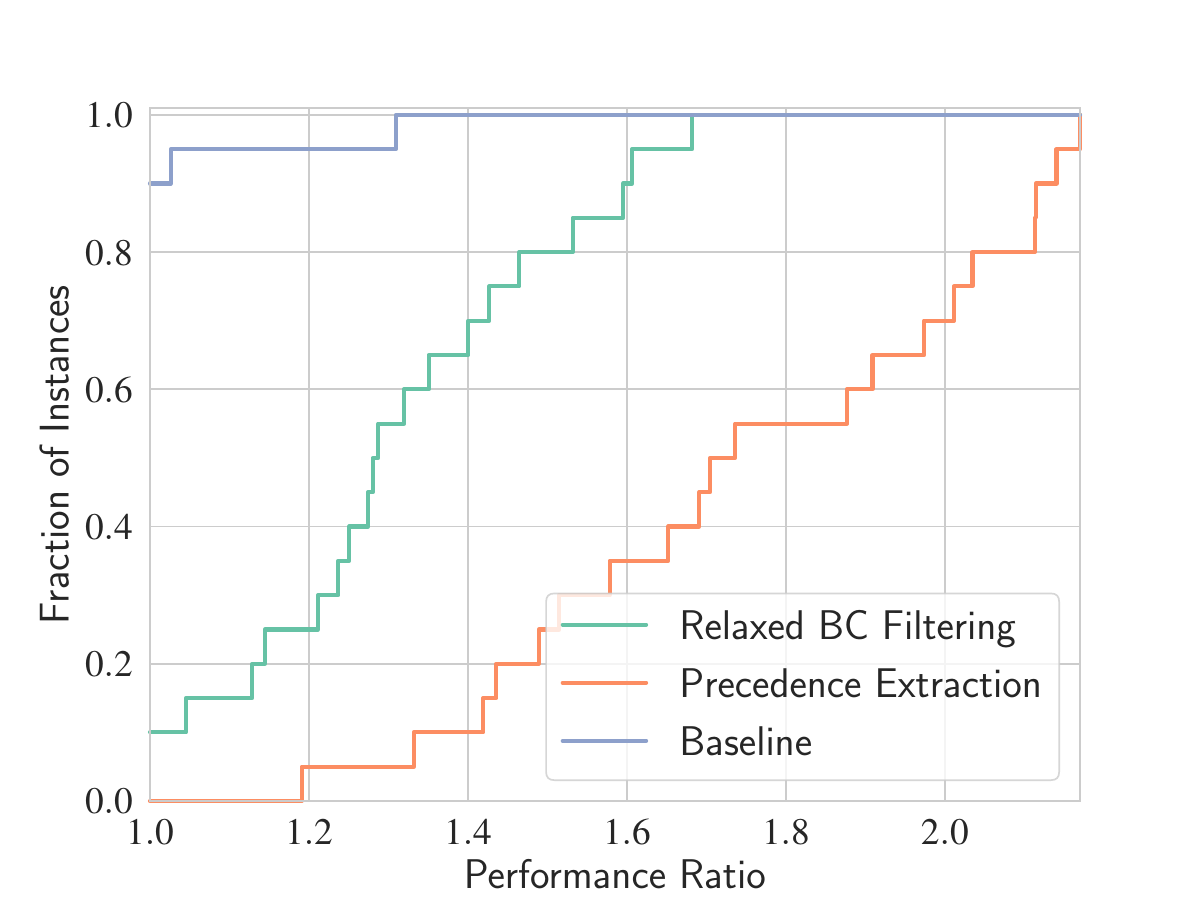} 
    \caption{Time Performance profile}
    \label{fig:ppTime}
\end{subfigure}

\begin{subfigure}{0.49\textwidth}
\centering
    \includegraphics[width=0.49\textwidth]{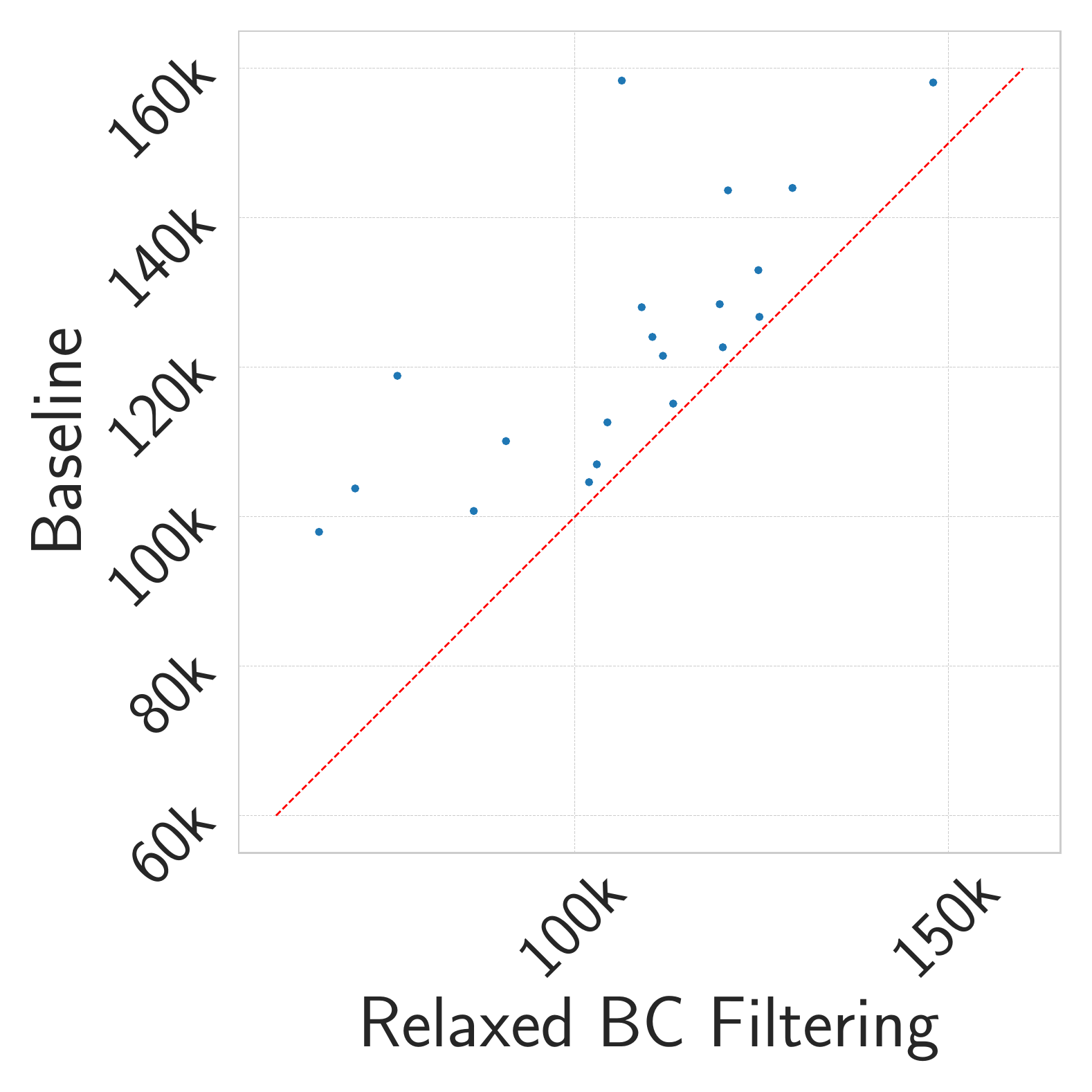} 
    \includegraphics[width=0.49\textwidth]{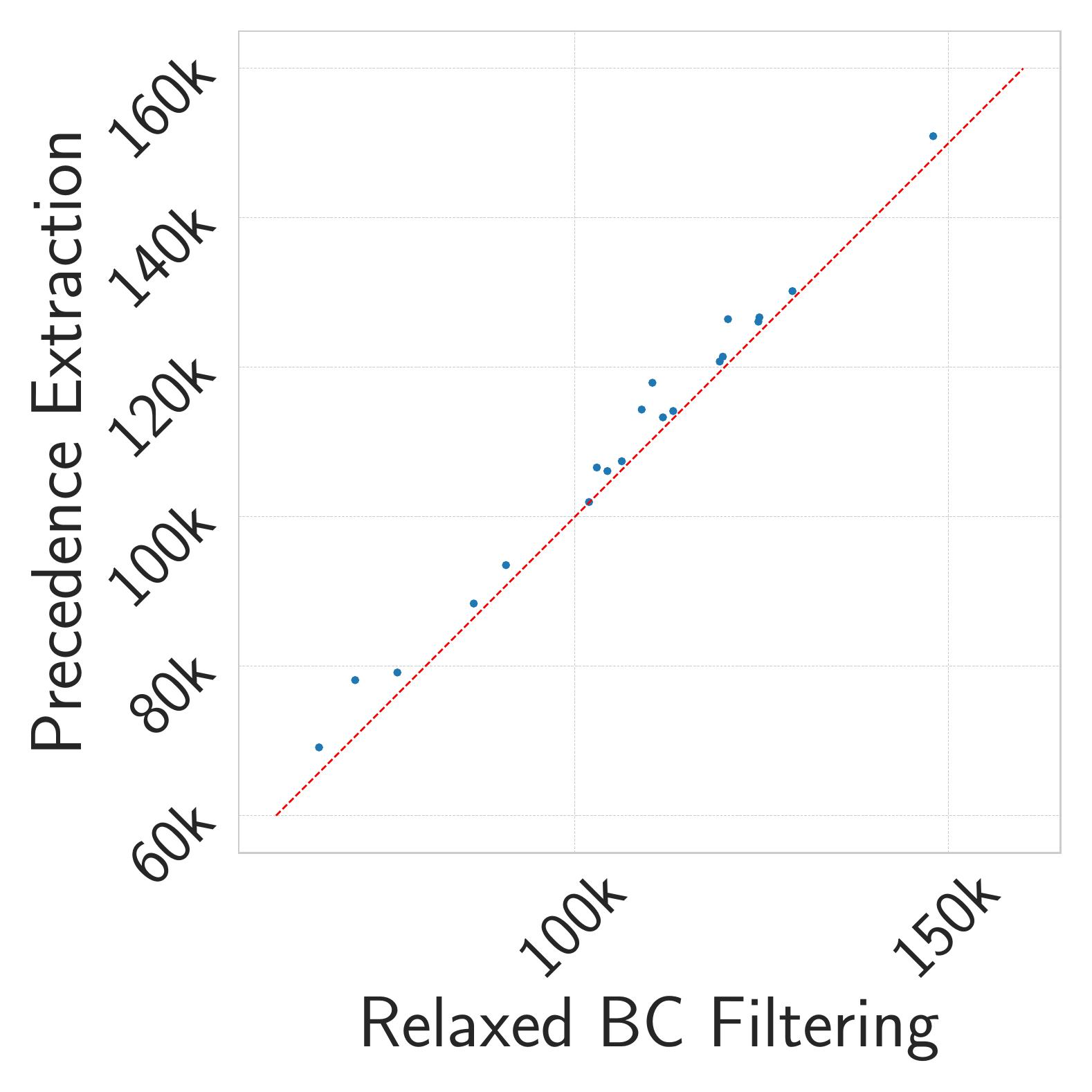} 
    \caption{Plot of the search node counts}
    \label{fig:scatterNodesComplete}
\end{subfigure}
    \caption{Results for instances of size $n=40$ and relaxed \mdds bounded to $W=16$}
    \label{fig:results}
\end{figure}
In this experiment, comparing the baseline, the model with relaxed BC filtering, and the model with precedence extraction, twenty instances of the problem were generated for each $n\in \{18, 25, 30, 40\}$. Fig.~\ref{fig:ppTime} displays a performance profile \cite{dolan2002benchmarking} (i.e., a cumulative over the instances of the ratio between the method's performance and the best performance overall) of the times of each method. The graph shows that our method outperforms the precedence extraction but is dominated by the baseline. Appendix contains the results for smaller instances.
We also empirically confirmed the quadratic time complexity nature of Relaxed BC Filtering, in contrast to the cubic one of precedence extraction, by comparing the time taken at each node of the search tree. Complete results are in the Appendix.

\mdd{}-based methods systematically reduce the number of nodes explored (Fig.\ref{fig:scatterNodesComplete}). In addition, the relaxed BC filtering is even more reducing the search tree than extracting precedence, showing that even in a relaxed context, the effect of the stronger propagation is shown (Fig.\ref{fig:scatterNodesComplete}). 

We also varied the \mdd size to assess the impact. The results in the Appendix are consistent with those of previous experiments in \cite{Cire2012,Gentzel2020}. The wider the \mdd, the greater the reduction in the search tree, but there is an increasing exponential overhead of the operations related to construction and refinement of the \mdd. A size for the \mdd of 16 has also been shown (as in \cite{Cire2012,Gentzel2020}) to be a good compromise between relaxation quality and time spent.

\subsection{Comparison to Bound-Consistency}

\begin{figure}[t!]
    \centering
    \includegraphics[width=\linewidth]{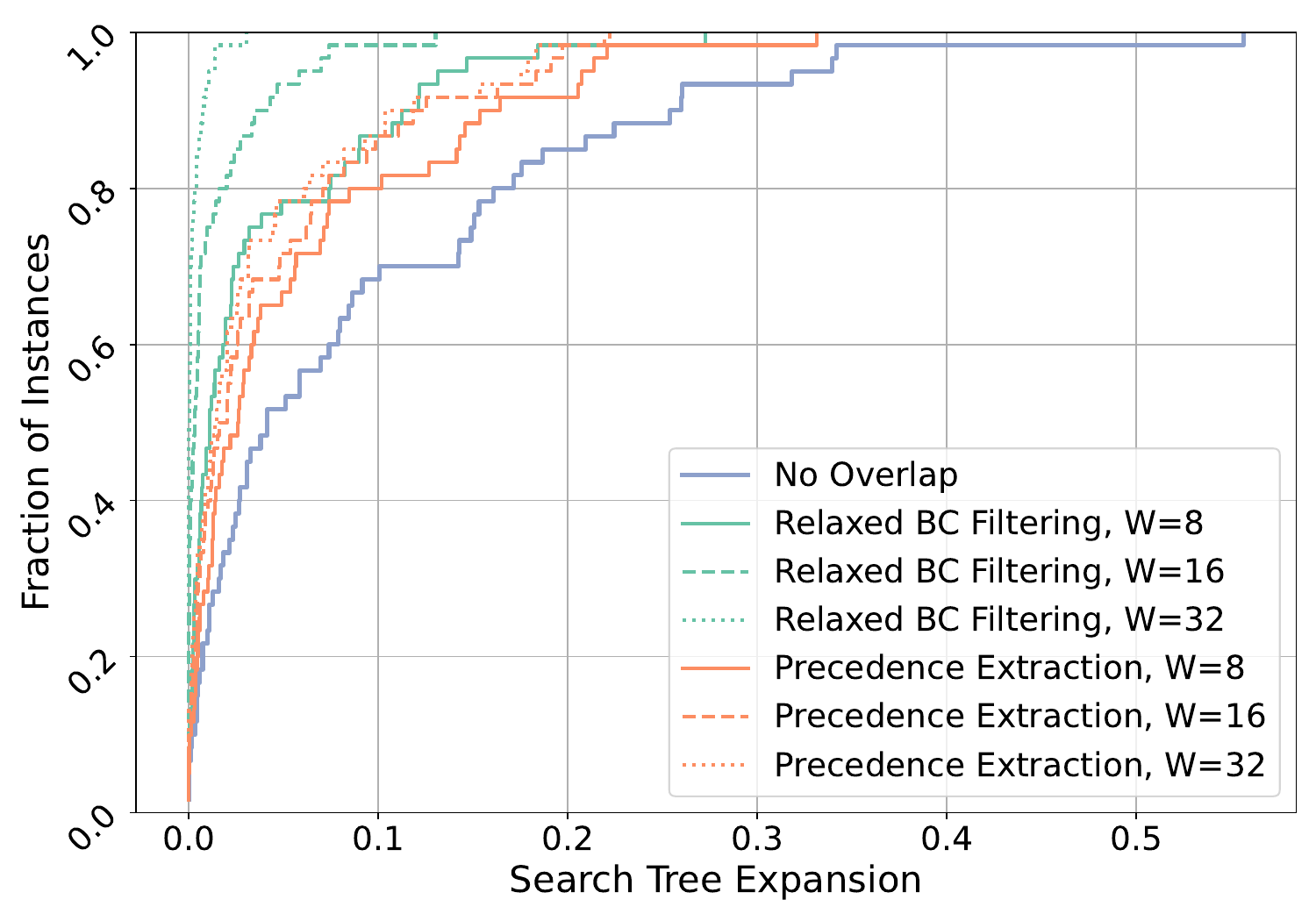}
    \caption{Cactus plot showing, for each method and MDD width, the proportion of replayed searches completed with a given gap $g$ relative to the exact bound-consistent method. Smaller search tree expansion indicate behavior closer to bound-consistency.}
    \label{fig:gap_cactus_plot}
\end{figure}

\begin{table}[t!]
{\footnotesize
\centering
\begin{tabular}{lcc||ccc}
\toprule
Method & $n$ & $W$ & min (s) & max (s) & mean (s)\\
\midrule
BC (exact \nomdd) & 12 & - & 4.6 & 449.0 & 82.7 \\
Baseline & 12 & - & 0.7 & \textbf{18.0} & 5.7  \\
Relaxed BC & 12 & 8 & 0.3 & 20.4 & 4.3  \\
Relaxed BC & 12 & 16 & \textbf{0.2} & 18.6 & \textbf{4.1}  \\
Relaxed BC & 12 & 32 & 0.3 & 20.4 & 4.3  \\
PE & 12 & 8 & \textbf{0.2} & 21.1 & 4.4  \\
PE & 12 & 16 & 0.3 & 21.1 & 4.2  \\
PE & 12 & 32 & 0.3 & 21.2 & 4.4  \\
\midrule
BC (exact \nomdd)  & 14 & - & 121.8 & 7141.5 & 2066.9  \\
Baseline & 14 & - & 4.9 & \textbf{60.0} & \textbf{24.1}  \\
Relaxed BC & 14 & 8 & 2.6 & 82.1 & 27.7  \\
Relaxed BC & 14 & 16 & 2.5 & 98.3 & 29.2  \\
Relaxed BC & 14 & 32 & 2.2 & 70.6 & 26.0  \\
PE & 14 & 8 & \textbf{2.1} & 91.6 & 30.4  \\
PE & 14 & 16 & \textbf{2.1} & 117.0 & 32.1  \\
PE & 14 & 32 & 2.2 & 89.2 & 29.9 \\
\midrule
BC (exact \nomdd)  & 16 & - & 1183.8 & 20630.4 & 8991.9  \\
Baseline & 16 & - & \textbf{2.7} & \textbf{60.0} & \textbf{27.4} \\
Relaxed BC & 16 & 8 & 3.2 & \textbf{60.0} & 28.1  \\
Relaxed BC & 16 & 16 & 3.7 & 64.1 & 28.5 \\
Relaxed BC & 16 & 32 & 3.7 & 61.8 & 27.7 \\
PE & 16 & 8 & 3.3 & 65.4 & 31.3 \\
PE & 16 & 16 & 4.0 & 75.6 & 32.6 \\
PE & 16 & 32 & 4.0 & 71.0 & 32.2 \\
\bottomrule
\end{tabular}
}
\caption{Comparison of search time in seconds for different methods, problem size, and \nomdd width.}
\label{tb:time}
\end{table}

Having a Bound-Consistent algorithm at our disposal, even if inefficient in practice for larger instances, allows to empirically compare, on smaller instances, the pruning capacity of weaker algorithms relative to Bound-Consistency (i.e., how many nodes are unnecessarily explored compared to BC). For this experiment, 20 instances of sizes 12, 14, and 16 were generated. The replay method first records the search tree of the baseline model using a 1-minute timeout.  The number of nodes explored, and the search time for the four models, as well as the width for the exact \nomdd were monitored. 

Figure~\ref{fig:gap_cactus_plot} presents a cactus plot for each non-BC method of the search tree expansion between the method and the BC filtering. The search tree expansion $g$ is computed using the following formula:
$$g = \frac{Z - Z'}{Z'}$$
where $Z$ is the number of nodes explored during the search by the method under consideration, and $Z'$ is the number of nodes explored when using the bound-consistent propagator. The tree expansion denotes the additional percentage of nodes required by the method compared to the BC search tree. As all the compared methods have a weaker consistency than BC, the tree expansions measured here are all positives. Closer to a zero tree expansion means closer to a bound consistent pruning.

The results, comparing the four models (baseline, relaxed BC filtering, precedence extraction, and BC filtering), show first that the classical relaxation used in the baseline is relatively far from BC, requiring at least 10\% more nodes on 30\% of the tested instances and with only 20\% of the instances close ($<$1\% tree expansion) to BC. The precedence extraction behaves a bit better, requiring at least 10\% more nodes for 10 to 20\% of the instance (depending on the W). The relaxed BC model shows empirically the best results, requiring at least 10\% more node for 0 to 7\% of the instance. With a width of 32, it even shows a maximum of 5\% tree expansion for the whole benchmark.
For both models with \nomdd, a larger width led to smaller tree expansion, demonstrating the expected improved quality of a less relaxed \mdd.

Although the instances considered are small, it was observed that the exact \nomdd already reaches an average width of 20 000 nodes during its construction. This significantly limits scalability, as larger or real-world instances would quickly exceed practical memory constraints, resulting in prohibitive search times. Relax \mdds allows avoiding such an explosion of required memory by bounding the width.

Table~\ref{tb:time} reports the minimum, maximum, and average replay times for each method. For the baseline, precedence extraction and relaxed BC, the values follows similar trends as for bigger instances (see Fig.\ref{fig:results}). For the model with the BC filtering, the results shows the drastic increase (several order of magnitude higher) of time resulting from the exponential behavior of the exact algorithm.
This demonstrates that, for a fraction of the search time and a fraction of the \nomdd width required, our method is able to reach performance close to bound-consistency.

To summarize the findings of this experiment, our relaxed BC algorithm exhibits the lowest search tree expansion among the relaxed algorithms (gap less than $5\%$ for $W=32$ on each instance), while maintaining control over memory usage. 

\section{Conclusion}
\label{sec:conclusion}


This paper proposes the first BC algorithm to tighten the time window of the jobs in the \nooverlap constraint based on the \nooverlap \mdd introduced in \cite{Cire2012}. A relaxed polynomial-time version of it is obtained by relaxing the \mdd, i.e., bounding its width. 
Experiments on a single-machine just-in-time scheduling problem showed that the new filtering method provides additional pruning compared to the standard filtering for the no-overlap constraint. The experiments also show that, compared to the classical filtering only, using the Relaxed BC filtering redundantly achieves filtering closer to bound-consistency.


\bibliographystyle{named}
\bibliography{bib}

\end{document}